**TITLE:** Social media mining for toxicovigilance of prescription medications: End-to-end pipeline, challenges and future work

**Authors:**

[a]Abeed Sarker, PhD (abeed.sarker@emory.edu)

**KEYWORDS:** substance use; substance use disorder; social media; natural language processing; data science

**Author affiliations:**

[a]Department of Biomedical Informatics, School of Medicine, Emory University, Woodruff Memorial Research Building, 101 Woodruff Circle, Suite 4101Atlanta, GA 30322, USA

**Corresponding author: Abeed Sarker**
101 Woodruff Circle, Suite 4101, Atlanta, GA 30032, USA
Email: abeed.sarker@emory.edu

**ABSTRACT**

Substance use, substance use disorder, and overdoses related to substance use are major public health problems globally and in the United States. A key aspect of addressing these problems from a public health standpoint is improved surveillance. Traditional surveillance systems are laggy, and social media are potentially useful sources of timely data. However, mining knowledge from social media is challenging, and requires the development of advanced artificial intelligence, specifically natural language processing (NLP) and machine learning methods. We developed a sophisticated end-to-end pipeline for mining information about nonmedical prescription medication use from social media, namely Twitter and Reddit. Our pipeline employs supervised machine learning and NLP for filtering out noise and characterizing the chatter. In this paper, we describe our end-to-end pipeline developed over four years. In addition to describing our data mining infrastructure, we discuss existing challenges in social media mining for toxicovigilance, and possible future research directions.

**Main Article**

Substance use, including nonmedical use of prescription medications, substance use disorder, and overdoses related to substance use are major public health problems in the United States (US) and globally. According to the latest estimates (available in November 2022), in the 12 months leading up to May 2022, more than 100,000 overdose-related deaths occurred in the US (over 275 deaths per day on average).[1] Despite many years of effort, it has thus far not been possible to curb the opioid epidemic in the US. While there are many reliable traditional sources of information, such as the WONDER database from the Centers for Disease Control and Prevention (CDC) for tracking overdose[2] deaths and the National Survey on Drug Use and Health (NSDUH),[3] these sources are often *laggy*. There are substantial lags associated with the process of data collection, curation, and publication. Overdose death data, for example, may take more than a year to compile and publish. Consequently, trends in substance use and its impact are only known retrospectively. Often, by the time we are able to obtain a complete picture of the trends within a given time period, considerable damage has already been done and the patterns have shifted. This is particularly problematic because patterns in population-level substance use are constantly evolving.[4] There is thus a need for establishing complementary sources and methods for surveillance. Social media, coupled with methods for mining knowledge from them, have the potential of serving as timely and complementary sources of information for substance use. However, there has been limited research on developing and validating methods for leveraging social media data for substance use surveillance or *toxicovigilance*. In this paper, we outline our progress in developing the social media mining infrastructure needed for mining substance use-related knowledge from social media. Our specific focus was prescription medications, including opioids, benzodiazepines, and stimulants. In the following paragraphs, we outline our methodological infrastructure, including natural language processing (NLP) and machine learning methods, and findings over a period of four years of research. Figure 1 provides an outline of the entire pipeline, using Twitter as an example.

*Data collection and annotation*

The first step in successfully leveraging social media data for toxicovigilance is to establish a data collection strategy. For our work, we initially focused on Twitter and later integrated data from Reddit. For Twitter, we collected data about a given set of prescription medications including opioids (such as oxycodone), stimulants (such as Adderall®), and benzodiazepines (such as alprazolam). We used the Twitter academic application programming interface (API) for collecting data using the medication names (generic and trade) as keywords. Since medication names are often misspelled by Twitter subscribers, we had to devise a strategy for incorporating common misspellings for our chosen set of medication names. We developed an automatic, data-centric misspelling generator that used a phrase embedding model learned from social media and a recursive algorithm that combined semantic and lexical similarity measures.[5] We found that the inclusion of misspellings automatically generated by our algorithm increased our post retrieval rate by over 30% and we later extended the

algorithm for generating multi-word lexical variants.[6]

In line with past literature,[7] a manual review of the posts retrieved by our data collection mechanism revealed that only a small portion of all posts represented personal nonmedical use, while most were simply mentions of the medications (*e.g.*, sharing of News articles). Since our focus was on studying nonmedical use of prescription medications, we decided against conducting unsupervised analysis of the entire data and instead decided to apply a supervised machine learning filter to automatically identify the posts that represented nonmedical use. As a first step, in the absence of annotated data for this supervised classification task, we prepared a detailed annotation guideline and then manually annotated data into four classes—(i) nonmedical use, (ii) consumption, (iii) mention, and (iv) unrelated. We annotated a total of 16,443 posts, obtaining an average, pairwise inter-annotator agreement of 0.86 (Cohen's kappa[8]). The annotated guideline and the dataset are available for research purposes.

Data collection from Reddit was substantially simpler since Reddit consists of many special-interest communities (subreddits) that host topic-specific chatter. For our targeted studies, we first identified subreddits of interest and then collected all posts available via the PRAW API.[9]

*Supervised classification*
Using the manually-annotated data, we experimented with several supervised learning algorithms to identify the best strategy. Specifically, we compared the performances of traditional classification models such as support vector machines (SVMs), deep learning methods, and transformer-based methods.[10] We found that fusion-based classifiers involving multiple transformer-based models achieved the best performance in terms of $F_1$ score (0.67) for the nonmedical use (minority) class. Because of the challenging nature of this classification task, we later attempted to improve the classification performance for the nonmedical use class by incorporating additional innovations, such as source-adaptive pretraining and topic-specific pretraining.[11] We also attempted to promote community-driven development of effective solutions for this task by proposing it in the social media mining for health applications (SMM4H) shared tasks, 2020.[12]

Our approach for detecting self-reports of nonmedical prescription medication use enabled us to create what is to date the largest social media based cohort for nonmedical prescription medication use. At the time of writing, this cohort consists of over 650,000 members. Each member of the cohort have been automatically detected at some point to have publicly expressed nonmedical prescription medication use. Using the API, we collected all publicly-available past posts of each cohort member, and we repeated this collection strategy every two weeks, resulting in multi-year longitudinal timelines of these cohort members. We used the classified posts and the longitudinal data for targeted, downstream analyses.

*Post-classification tasks*
We conducted several studies in which we tried to further filter our data to remove noise. For example, we developed methods for detecting and removing bots from our cohort,[13] comparing therapeutic and recreational use of opioids from Twitter data by employing a multi-class classification strategy,[14] and automating the detection of illicit opioid use.[15] For some of our targeted analysis of Twitter data, we used only post-level data samples (*i.e.*, only the posts that contained the medication names rather than longitudinal data from the cohort members). Such studies were particularly conducted early on in our project when sufficient amounts of longitudinal cohort data had not been collected. For example, we conducted thematic analyses to study provider perceptions about buprenorphine initiation,[16] and compare chatter regarding medications for opioid use disorder such as buprenorphine-naloxone and methadone.[17]

A key advance made by our Twitter-based approach was in geolocation-centric analysis. As mentioned earlier, one of our key motivations was to detect signals of nonmedical use from social media earlier than other sources. Hence, we wanted to compare past signals that we could collect from social media with known metrics from traditional sources, such as the CDC WONDER database and NSDUH. Using the state of Pennsylvania as our subject, we compared the rates of tweets classified as nonmedical use with overdose death numbers at the county level and several relevant metrics from the NSDUH at the substate level.[18] We found significant correlations between the social media estimates and the metrics from the traditional sources, suggesting that social media data may serve as a complementary resource for predicting substance use and related overdose deaths.

Our more recent work focused on leveraging the longitudinal data posted by our cohort members. In particular, we attempted to address one key limitation of social media data compared to the NSDUH and other traditional sources of information. The latter typically have demographic information available (*e.g.*, biological sex/gender identify and race), which social media data did not have. Therefore, we developed methods for automatically estimating the distribution of gender identities, race, and age groups from our cohort data and compared them to those reported in traditional sources. Broadly speaking, we found the binary gender distributions to largely agree with traditional sources,[19] often agreeing more with one source compared to another. For race we found strong correlation between estimates derived from Twitter and NSDUH.[20] We found moderate correlation for age-group-related metrics, which was explainable since the subscriber base of Twitter is skewed towards overrepresenting younger people and underrepresenting younger people. We also conducted a large-scale analysis of emotions expressed along with nonmedical prescription medication use, and we compared these between cohort subsets.[21] Our analyses revealed significant differences between the emotions and concerns expressed by people who nonmedically use stimulants, opioids and benzodiazepines compared to those who do not. Our analyses also revealed significant differences between the emotional contents of men and women who report nonmedical use of prescription medications.

*Reddit data analysis*
While we used Twitter primarily to progress the public health aspect of our work, we found Reddit to be particularly helpful in providing us with clinical insights. Early on in our project, we used Reddit to understand consumer perceptions about medications for opioid use disorder, such as buprenorphine-naloxone.[22] Unlike Twitter, we found Reddit to encapsulate intricate details about user experiences and perceptions. This is particularly the case because Reddit does not impose length limits at the post level, unlike Twitter, and the platform is built on the concept of anonymity (*i.e.*, subscribers can remain anonymous if they desire). Consequently, discussions on Reddit are often candid and contain more depth. In a later study, we utilized data from Reddit to understand the experiences of people who use opioids in terms of precipitated withdrawal when initiating buprenorphine treatment. We found that while the literature lacked information about this problem that is being observed relatively commonly in emergency departments, Reddit subscribers had been discussing this topic for multiple years. We even found that Reddit subscribers had advocated self-management strategies for avoiding precipitated withdrawal, including using a specific microdosing strategy called the *Bernese method*.[23]

*Conclusion and future work*

Our deployed pipeline will continue to automatically collect cohort members based on their self-reported nonmedical prescription medication use. Longitudinal data from cohort members will also continue to be collected. Thus, our cohort dataset is the largest of its kind and it will uniquely preserve longitudinal data posted by the cohort members. While our completed work represents considerable progress in this research space, there are a number of important future directions to expand our work. We outline some of these future research tasks below.

i. Expanding to illicit substances: this is a natural extension of our current pipeline which focuses on nonmedical use of prescription medications only. We plan to incorporate illicit substances into our pipeline. This will require manual annotation of additional data and training new supervised classification models. Note that there is a subtle difference in the classification process between prescription and illicit substances. Any consumption of illicit substances is considered nonmedical use, but that is not the case for prescription medications.

ii. Discovering novel psychoactive substances: this is an area where social media may have a high impact. Emerging and novel substances often appear on social media first before they are detected by traditional systems. Our preliminary work in this space focusing on benzodiazepines suggests that it may be possible to obtain strong signals about novel benzodiazepines before they become widespread in the population.[24]

iii. Studying long-term impacts and trajectories of substance use: our massive and growing cohort will have multiple years of longitudinal data available. These can be used to study long-term trajectories and impacts. While impacts, such as social and clinical impacts, can be sparsely occurring, the availability of a large cohort may make the detection of patterns possible. Our preliminary works in analyzing trends have produced promising results, and we envision the application of few-/low-shot learning methods for automatically detecting sparse concepts.
iv. Creating a publicly available resource: while most of our data and methods are public, there is the potential to make aggregated statistics available to the broader research community in easy-to-use formats. For example, creating a web-based dashboard that enables the easy download of aggregated statistics may help public health and related researchers.
v. Reporting back to the community: while we conduct research using publicly available data, we rarely report back to the people whose data are being used in the research. Our vision is to establish a protocol for reporting the findings of our work back to the research community.

**Funding**

Research reported in this publication is supported by the NIDA of the NIH under the award numbers R01DA046619 and R01DA057599. The content is solely the responsibility of the authors and does not necessarily represent the official views of the NIH.

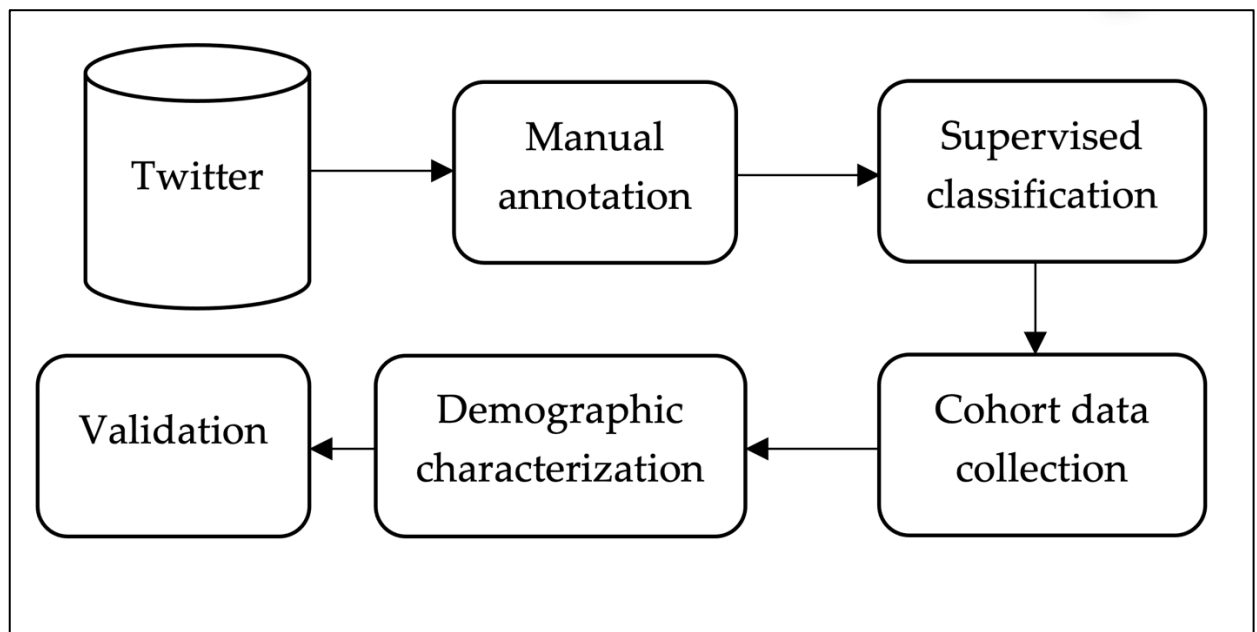

**Figure 1.** End-to-end pipeline for toxicovigilance research from Twitter data.